\documentclass[journal]{IEEEtran}

\usepackage[utf8]{inputenc}
\usepackage{amsmath}
\usepackage{amsfonts,amssymb}
\usepackage{url}
\usepackage{graphicx}

\usepackage{float}
\usepackage{multirow}

\usepackage{tikz}
\usepackage{pgf}
\usepackage{placeins}

\usepackage{comment}
\usepackage{url}
\usepackage{placeins}
\usepackage{latexsym}
\usepackage{pifont}
\usepackage{pgfplots,filecontents}
\usepackage{gensymb}
\usepackage{url}
\usepackage{caption}
\usepackage{algorithm}
\usepackage{algorithmic}

\pgfplotsset{compat=newest}

\usetikzlibrary{intersections}
\usetikzlibrary{positioning,angles,quotes,patterns}

\newcommand{\R}{\mathbb{R}}

\newcommand{\E}{\mathbb{E}}
\newcommand{\ra}{\rightarrow}

\newtheorem{assumption}{Assumption}

\DeclareGraphicsExtensions{.png,.jpg,.jpeg,.eps,.gif}
\tikzstyle{block} = [rectangle, draw,text width=3em, text centered, rounded corners]

\tikzstyle{block2} = [rectangle, draw, text centered, rounded corners]
    
\tikzstyle{line} = [draw, -latex]

\begin{document}
\nocite{*}
\title{A Survey of Reinforcement Learning Algorithms for Dynamically Varying Environments}
%

\author{Sindhu~Padakandla \thanks{Sindhu P R is with the Dept. of Computer Science and Automation, Indian Institute of Science, Bangalore, Karnataka, 560012 India E-mail: sindhupr@iisc.ac.in.}}
\maketitle

\begin{abstract}
Reinforcement learning (RL) algorithms find applications in inventory control, recommender systems, vehicular traffic management, cloud computing and robotics. The real-world complications of many tasks arising in these domains makes them difficult to solve 
with the basic assumptions underlying classical RL algorithms.
RL agents in these applications often need to react and adapt to changing operating conditions.
A significant part of research on single-agent RL techniques focuses on developing algorithms when the underlying assumption 
of stationary environment model is relaxed.
This paper provides a survey of RL methods developed for handling dynamically varying environment models. The goal of methods not limited by the stationarity assumption is to help autonomous agents adapt to varying operating conditions. This is possible either by minimizing the rewards lost during learning by RL agent or by finding a suitable policy for the RL agent which leads to efficient operation of the underlying system.
A representative collection of these algorithms is discussed in detail in this work along with their categorization and their relative merits and demerits. Additionally we also review works which are tailored to application domains. Finally, we discuss future enhancements for this field. 

\end{abstract}

\section{Introduction}\label{sec:intro}
Resurgence of artificial intelligence (AI) and advancements in it has led to automation of physical and cyber-physical systems~\cite{rlincps}, cloud computing~\cite{rlincloud}, communication networks~\cite{rlincomm}, robotics~\cite{rlinrobotics} etc. Intelligent automation through AI requires that
these systems be controlled by smart \emph{autonomous agents} with least manual intervention.
Many of the tasks in the above listed applications are of \emph{sequential decision-making} nature, in the sense that the autonomous agent monitors the \emph{state} of the system and decides on an \emph{action} for that state. This action when exercised on the system,  \emph{changes} the state of the system. Further, in the new state, the agent again needs to choose an action (or control). This repeated interaction between the autonomous agent and the system is sequential and the change in state of the system is dependent on the action chosen. However, this change is uncertain and the future state of the system cannot be predicted.
For e.g., a recommender system~\cite{recommender} controlled by an autonomous agent seeks to predict ``rating" or ``preference" of users for commercial items/movies. Based on the prediction, it recommends items-to-buy/videos-to-watch to the user. Recommender systems are popular on online stores, video-on-demand service providers etc.. In a recommender application, the \emph{state} is current genre of videos watched or books purchased etc., and the agent decides on the set of items to be recommended for the user. Based on this, the user chooses the recommended content or just ignores it. After ignoring the recommendation, the user may go ahead and browse some more content. In this manner, the state \emph{evolves} and every action chosen by the agent captures additional information about the user.

It is important to understand that there must be a \emph{feedback} mechanism which recognizes when the autonomous agent has chosen the \emph{right} action. Only then can the autonomous agent \emph{learn} to select the right actions.
This is achieved through a \emph{reward (or cost)} function which ranks an action selected in a particular state of the system.
Since the agent's interaction with the system (or \emph{environment}) produces a sequence of actions, this sequence is also ranked by a pre-fixed \emph{performance criterion}. Such a criterion is usually a function of the rewards (or cost) obtained throughout the interaction.
The goal of the autonomous agent is to find a sequence of actions for every initial state of the system such that this performance criterion is optimized in an average sense. Reinforcement learning (RL)~\cite{sutton} algorithms provide a mathematical framework for sequential decision making by autonomous agents.

In this paper, we consider an important challenge for developing autonomous agents for real-life applications~\cite{arxiv2019}.
This challenge is concerned with the scenario when the environment undergoes changes. Such changes necessitate 
that the autonomous agent continually track the environment characteristics and adapt/change the learnt actions in order to ensure efficient system operation. For e.g., consider a vehicular traffic signal junction managed by an autonomous agent. This is an example of intelligent transportation system, wherein the agent selects the green signal duration for every lane. The traffic inflow rate on lanes varies according to time of day, special events in a city etc. 
If we consider the lane occupation levels as the \emph{state}, then the lane occupation levels are influenced by traffic inflow rate as well as the number of vehicles allowed to clear the junction based on the green signal duration. 
Thus, based on traffic inflow rate, some particular lane occupation levels will be more probable. 
If this inflow rate varies, some other set of lane occupation levels will become more probable. Thus, as this rate varies, so does the state evolution distribution. 
It is important that under such conditions, the agent select appropriate green signal duration based on the traffic pattern and it must be adaptive enough to change the selection based on varying traffic conditions.

\begin{figure}[ht]
\begin{center}
%
\includegraphics[scale=0.62]{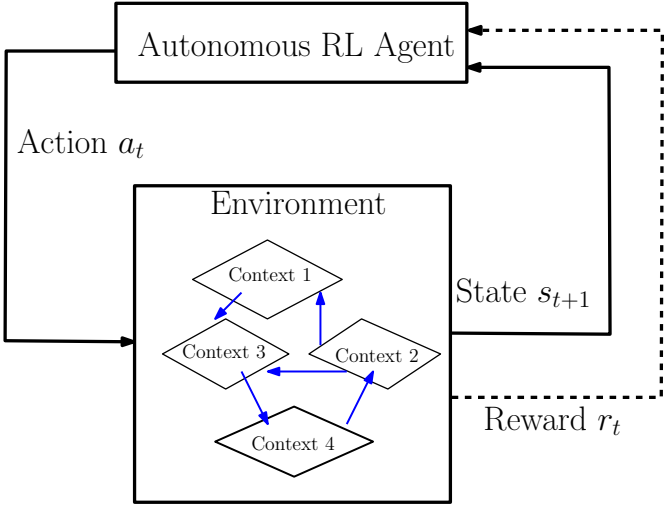}
\end{center}

\caption{Reinforcement learning with dynamically varying environments. The environment is modeled as a set of contexts and evolution of these contexts is indicated by blue arrows. At time $t$, the current state is $s_t$ and RL agent's action $a_t$ changes the state to $s_{t+1}$ and reward $r_t$ is generated.} 
\label{fig:nonstatrl}
\end{figure}

Formally, the system or environment is characterized by a \emph{model} or \emph{context}. The model or context comprises of the state evolution probability distribution and the reward function - the first component models the uncertainty in state evolution, while 
the second component helps the agent learn the right sequence of actions. 
The problem of varying environments implies that the environment context changes with time. This is illustrated in Fig. \ref{fig:nonstatrl}, where the environment model chooses the reward and next state based on the current ``active'' context $i$, $1 \leq i \leq n$. More formal notation is described in Section \ref{sec:prelim}.

\subsection{Contribution and Related Work}
\label{subsec:contrib}
This paper provides a detailed discussion of the reinforcement learning techniques for tackling dynamically changing environment contexts in a system. The focus is on a single autonomous RL agent learning a sequence of actions for controlling such a system. 
Additionally, we provide an overview of challenges and benefits of developing new algorithms for dynamically changing environments. 
The benefits of such an endeavour is highlighted in the application domains where the effect of varying environments is clearly observed. We identify directions for future research as well.

Many streams of work in current RL literature attempt to solve the same underlying problem - that of learning policies which ensure proper and efficient system operation in case of dynamically varying environments. This problem will be formally defined in Section \ref{sec:probdef}. However, here we give an overview of the following streams of work : \emph{continual learning} and \emph{meta-learning}. In Section \ref{sec:relatedwork}, a detailed analysis of prior works in these streams is provided.
\begin{itemize}
 
 \item \emph{Continual learning}: Continual learning~\cite{clsurvey} is the ability of a model to learn continually from a stream of data, building on what was learnt previously, as well as being able to remember previously seen tasks. The desired properties of a continual learning algorithm is that it must be able to learn every moment, transfer learning from previously seen data/tasks to new tasks and must be resistant to \emph{catastrophic forgetting}. Catastrophic forgetting refers to the situation where the learning algorithm forgets the policies learnt on previous models while encountering new environment models. Recent continual learning algorithms in machine learning and neural networks is covered in \cite{clsurvey}.
 
 \item \emph{Meta-learning}: Meta-learning~\cite{mlsurvey} or ``learning to learn'' involves observing how a learning algorithm performs on a wide range of tasks and then using this experience to learn new tasks quickly in a much more efficient and skilled manner. Thus, meta-learning emphasizes on putting experience to use and reusing approaches that worked well before. With advancements in neural network architectures, there is a renewed interest in improving meta-learning algorithms and \cite{mlsurvey} reviews prior works.
\end{itemize}

\subsection{Overview}
The remainder of the paper is organized as follows. Section \ref{sec:prelim} presents the basic mathematical foundation for modelling a sequential decision-making problem in the Markov decision process (MDP) framework. It also briefly states the assumptions which are building blocks of RL algorithms.
In Section \ref{sec:probdef}, we formally introduce the problem, provide a rigorous problem statement and the associated notation. 
Section \ref{sec:ben-cha} describes the benefits of developing algorithms for dynamically varying environments. It also identifies challenges that lie in this pursuit.
Section \ref{sec:relatedwork} describes the solution approaches proposed till now for the problem described in Section \ref{sec:probdef}. This section discusses two prominent categories of prior works. Section \ref{sec:relatedareas} discusses relevant works in continual and meta learning. In both Sections \ref{sec:relatedwork} and \ref{sec:relatedareas}, we identify the strengths of the different works as well as the aspects that they do not address. 
Section \ref{sec:appln} gives a brief overview of application domains which have been specifically targeted by some authors. 
Section \ref{sec:fw} concludes the work and elaborates on the possible future enhancements with respect to the prior work. Additionally, it also describes challenges that research in this area should address.

\section{Preliminaries}\label{sec:prelim}
Reinforcement learning (RL) algorithms are based on a stochastic modelling framework known as Markov decision process (MDP)~\cite{BertB,puterman}. In this section, we describe in detail the MDP framework.
\subsection{Markov Decision Process : A Stochastic Model}\label{subsec:mdp}
A MDP is formally defined as a tuple $M = \langle S, A, P, R\rangle$, where $S$ is the set of states of the system, $A$ is the set of actions (or decisions). $P:S \times A \ra \mathcal{P}(S)$ is the conditional transition probability function. 
Here, $\mathcal{P}(S)$ is the set of probability distributions over the state space $S$.
The transition function $P$ models the uncertainty in the evolution of states of the system based on the action exercised by the agent. Given the current state $s$ and the action $a$, the system evolves to the next state according to the probability distribution $P(\cdot|s,a)$ over the set $S$. 
At every state, the agent selects a feasible action for every \emph{decision epoch}. 
The decision horizon is determined by the number of decision epochs. 
If the number of decision epochs is finite (or infinite), the stochastic process is referred to as a \emph{finite (or infinite)-horizon} MDP respectively. 
$R : S \times A \ra \R$ is the reward (or cost) function which helps the agent learn.
The environment context comprises of the transition probability and reward functions. If environments vary, they share the state and action spaces but differ only in these functions.

\subsection{Decision Rules and Policies}\label{subsec:policy}
The evolution of states, based on actions selected by agent until time $t$, is captured by the ``history" variable $h_t$.
This is an element in the set $H_t$, which is the set of all plausible histories upto time $t$. 
Thus, $H_t = \{h_t = (s_0,a_0,s_1,a_1,\ldots,s_t) : s_i \in S, a_i \in A , \, 0 \leq i \leq t \}$. 
The sequence of decisions taken by agent is referred to as \emph{policy}, wherein a policy is comprised of \emph{decision rules}.
A randomized, history-dependent decision rule at time $t$ is defined as $u_t : H_t \ra \mathcal{P}(A)$, where $\mathcal{P}(A)$ is the set of all probability distributions on $A$.  Given $u_t$, the next action at current state $s_t$ is picked by sampling an action from the probability distribution $u_{t}(h_t)$. 
If this probability distribution is a degenerate distribution, then the decision rule is called \emph{deterministic decision rule}. 
Additionally, if the decision rule does not vary with time $t$, we refer to the rule as a \emph{stationary decision rule}. 
A decision rule at time $t$ dependent only on the current state $s_t$ is known as a state-dependent decision rule and denoted as $d_t:S \ra \mathcal{P}(A)$. 
A deterministic, state-dependent and stationary decision rule is denoted as $d: S \ra A$. 
Such a rule maps a state to its feasible actions. 
When the agent learns to make decisions, basically it learns the appropriate decision rule for every decision epoch.
A policy is formally defined as a sequence of decision rules. Type of policy depends on the common type of its constituent
decision rules.

\subsection{Value Function : Performance Measure of a Policy}\label{subsec:vfunc}
Each policy is assigned a ``score" based on a pre-fixed performance criterion (as explained in Section \ref{sec:intro}). 
For ease of exposition, we consider state-dependent deterministic decision rules only. 
For a finite-horizon MDP with horizon $T$, the often used performance measure is the expected total reward criterion. 
Let $\pi: S \ra A$ be a deterministic policy such that for a state $s$, $\pi(s) = (d_1(s),\ldots,d_T(s)), \, \forall s \in S$. 
The value function of a state $s$ with respect to this policy is defined as follows:
\begin{equation}
    \label{eqn:vpif}
 V^{\pi}(s) =  \E{\left[\sum\limits_{t=0}^{T} R(s_t,d_t(s_t)) | s_0 = s\right]},
\end{equation}
where the expectation is w.r.t all sample paths under policy $\pi$.
A policy $\pi^*$ is optimal w.r.t the expected total reward criterion if it maximizes \eqref{eqn:vpif} for all states and over all policies.

For infinite horizon MDP, the often used performance measures are the expected sum of discounted rewards of a policy and the average reward per step for a given policy.
Under the expected sum of discounted rewards criterion, 
the value function of a state $s$ under a given policy $\pi = (d_1,d_2,\ldots)$ is defined as follows:
\begin{equation}
\label{eqn:vpid}
 V^{\pi}(s) =  \E{\left[\sum\limits_{t=0}^{\infty} \gamma^t R(s_t,d_t(s_t)) | s_0 = s\right]}.
\end{equation}
Here, $0 \leq \gamma < 1$ is the discount factor and it measures the current value of a unit reward that is received one epoch in the future.
A policy $\pi^*$ is optimal w.r.t this criterion if it maximizes \eqref{eqn:vpid}.
Under the average reward per step criterion, the value function of a state $s$ under a given policy $\pi = (d_1,d_2,\ldots)$ is defined as follows (if it exists):
\begin{equation}
\label{eqn:vpia}
 V^{\pi}(s) =  \lim_{N\ra \infty} \frac{1}{N} \E{ \left[\sum\limits_{t=0}^{N-1} R(s_t,d_t(s_t)) | s_0 = s\right]}.
\end{equation}
The goal of the autonomous agent (as explained in Section \ref{sec:intro}) is to find a policy $\pi^*$ such that either \eqref{eqn:vpid} or \eqref{eqn:vpia} is maximized in case of infinite horizon or \eqref{eqn:vpif} in case of finite horizon, for all $s \in S$.

\subsection{Algorithms and their Assumptions}
RL algorithms are developed with basic underlying assumptions on the transition probability and reward functions. 
Such assumptions are necessary, since, RL algorithms are examples of \emph{stochastic approximation}~\cite{book:sa} algorithms. Convergence of the RL algorithms to the optimal value functions hold when the following assumptions are satisfied.
\begin{assumption}\label{as:rbdd}
$|R(s,a)| < \mathcal{B} < \infty$, $\forall a \in A \; \forall s \in S$.
\end{assumption}%
\begin{assumption}\label{as:statpr}
 Stationary $P$ and $R$, i.e., the functions $P$ and $R$ do not vary over time.
\end{assumption}%
Assumption \ref{as:rbdd} states that the reward values are bounded. Assumption \ref{as:statpr} implies that the transition probability and reward functions do not vary with time. 

We focus on model-based and model-free RL algorithmsin this survey. Model-based RL algorithms are developed to learn optimal policies and optimal value functions by estimating $P$ and $R$ from state and reward samples. Model-free algorithms do not estimate $P$ and $R$ functions. Instead these directly either find value function of a policy and improve or directly find the optimal value function. RL algorithms utilize \emph{function approximation} to approximate either the value function of a policy or the optimal value function. Function approximation is also utilized in the policy space. Deep neural network architectures are also a form of function approximation for RL algorithms~\cite{book:drl}.

In this paper, we use the terms ``dynamically varying environments'' and ``non-stationary environments'' interchangeably.
In the non-stationary environment scenario, Assumption \ref{as:statpr} does not hold true. Since previously proposed RL algorithms~\cite{sutton,ndp} are mainly suited for stationary environments, we need to develop new methods which autonomous agents can utilize to handle non-stationary environments.
In the next section, we formally describe the problem of non-stationary environments using the notation defined in this section. Additionally, we also highlight the performance criterion commonly used in prior works for addressing learning capabilities in dynamically varying environments.

\section{Problem Formulation}\label{sec:probdef}
In this section, we formulate the problem of learning optimal policies in non-stationary RL environments and 
introduce the notation that will be used in the rest of the paper. 
Since the basic stochastic modeling framework of RL is MDP, we will describe the problem using notation introduced in Section \ref{sec:prelim}.

We define a family of MDPs as $\{M_k\}_{k \in \mathrm{N}^+}$, where $M_k = \langle S, A, P_k , R_k \rangle$, 
where $S$ and $A$ are the state and action spaces, while $P_k$ is the conditional transition probability kernel and $R_k$ is the reward function of MDP $M_k$.
The autonomous RL agent observes a sequence of states $\{s_t\}_{t \geq 0}$, where $s_t \in S$. For each state, an action $a_t$ is chosen based on a policy. 
For each pair $(s_t,a_t)$, the next state $s_{t+1}$ is observed according to the distribution $P_k(\cdot | s_t,a_t)$ and reward $R_k(s_t,a_t)$ is obtained. Here $0 < k \leq t$. 
Note that, when Assumption \ref{as:statpr} is true, $P_k(\cdot | s_t,a_t) = P(\cdot | s_t,a_t), \; \forall k \in \mathrm{N}^+$ (as in Section \ref{sec:prelim}).
The RL agent must learn optimal behaviour when the system is modeled as a family of MDPs $\{M_k\}_{k \in \mathrm{N}^+}$. 

The decision epoch at which the environment model/context changes is known as \emph{changepoint} and we denote the set of changepoints using the notation $\{T_i\}_{i \geq 1}$, which is an increasing sequence of random integers. Thus, for example, at time $T_1$, the environment model will change from say $M_{k_0}$ to $M_{k_1}$, at $T_2$ it will change from $M_{k_1}$ to say
$M_{k_2}$ and so on. With respect to these model changes, the non-stationary \emph{dynamics} for $t \geq 0$ will be 
\begin{gather}\label{eqn:nonstatP}
P(s_{t+1} = s'|s_t = s,a_t = a) =    
\begin{cases}
  P_{k_0}(s'|s,a),\,\,  t < T_1\\    
  P_{k_1}(s'|s,a),\,\,  T_1 \leq t < T_2\\ 
  \vdots\\
\end{cases}
\end{gather}
and the reward for $(s_t,a_t) = (s,a)$ will be
\begin{gather}\label{eqn:nonstatR}
R(s,a) = 
\begin{cases}
  R_{k_0}(s,a),\,\,\, t < T_1\\    
  R_{k_1}(s,a),\,\,\, T_1 \leq t < T_2\\    
  \vdots\\
\end{cases}
\end{gather}
The extreme cases of the above formulation occur when either $T_{i+1} = T_{i} +1 , \; \forall i \geq 1$ or $T_1 = \infty$.
The former represents a scenario where model dynamics change in every epoch. The latter is the stationary case. Thus, the above formulation is a generalization of MDPs as defined in Section \ref{sec:prelim}.
Depending on the decision making horizon, the number of such changes will be either finite or infinite. 
With changes in context, the performance criterion differs, but \eqref{eqn:vpif}-\eqref{eqn:vpia} give away some hints as to what they can be. 
Additionally, since Assumption \ref{as:statpr} does not hold true, it is natural to expect that a stationary policy may not be optimal. Hence, it is important to expand the policy search space to the set of all history-dependent, randomized time-varying policies. 

Given the family of MDPs $\{M_k\}_{k \in \mathrm{N}^+}$, one \emph{objective} is to learn a policy $\pi = (u_1,u_2,\ldots)$ such that the long-run expected sum of discounted rewards, i.e., $\E{ \left[ \sum\limits_{t=0}^{\infty} \gamma^t R(s_t,u_t(H_t)) | H_0 = h_0 \right]}$ is maximized for all initial histories $h_0 \in H_0$. For finite horizon MDPs, the objective equivalent to \eqref{eqn:vpif} can be stated in a similar fashion. The same follows for \eqref{eqn:vpia}, where the policy search space will be randomized, history-dependent and time-varying. Another performance criterion which is widely used is called as \emph{regret}. This performance criterion is directly concerned with the rewards gained during system evolution, i.e., its more emphasis is on the rewards collected rather than  on finding the policy which optimally controls a system. The regret is usually defined for a finite-horizon system as follows:
\begin{equation}
    \label{eqn:regretdef}
    \textnormal{Regret} = V^*_T(s_0) - \sum_{t=0}^{T-1} R(s_t,a_t),
\end{equation}
where $T$ is the time horizon, $V^*_T(s_0)$ is the optimal expected $T$-step reward that can be achieved by any policy when system starts in state $s_0$.

It should be noted that the space of history-dependent, randomized policies is a large intractable space. Searching this space for a suitable policy is hard. Additionally, in the model-free RL case, how do we learn value functions with only state and reward samples? 
%
In the next section, we explore these issues and discuss prior approaches in connection with the problem of non-stationary environments in RL.
Some are methods designed for the case when model-information is known, while others are based on model-free RL. All regret-based approaches usually are model-based RL approaches, which work with finite-horizon systems. Approaches based on infinite-horizon systems usually are control methods, i.e., the main aim in such works is to find an approximately optimal policy for a system exposed to changing environment parameters. 

\section{Benefits and Challenges of RL in Non-Stationary Environments}\label{sec:ben-cha}
In this section we will indicate what are the benefits of tackling non-stationary environments in RL algorithms. These benefits straddle across single-agent and multi-agent scenarios. 
\subsection{Benefits}
RL is a machine learning paradigm which is more similar to human intelligence, compared to supervised and unsupervised learning. This is because, unlike supervised learning, the RL autonomous agent is not given samples indicating \emph{what} classifies as good behaviour and what is not. Instead the environment only gives a feedback recognizing \emph{when} the action by the agent is good and when it is not. Making RL algorithms efficient is the first step towards realizing general artificial intelligence~\cite{agi}. Dealing with ever-changing environment dynamics is the next step in this progression, eliminating the drawback that RL algorithms are applicable only in domains with low risk, for e.g., video games~\cite{rl_in_games} and pricing~\cite{pricing}.

Multi-agent RL~\cite{busoniu} is concerned with learning in prescence of multiple agents. It can be considered as an extension of single-agent RL, but encompasses unique problem formulation that draws from game theoretical concepts as well.
When multiple agents learn, they can be either competitive to achieve conflicting goals or cooperative to achieve a common goal. In either case, the agent actions are no longer seen in isolation, when compared to single-agent RL. Instead the actions are ranked based on what effect an individual agent's action has on the collective decision making. This implies that the dynamics observed by an individual agent changes based on other agents' learning. So, as agents continually learn, they face dynamically varying environments, where the environments are in this case dependent on joint actions of all agents. Unlike the change in transition probability and reward functions (Section \ref{sec:probdef}), when multiple agents learn, the varying conditions is a result of different regions of state-action space being explored. Thus, non-stationary RL methods developed for single-agent RL can be extended to multi-agent RL as well.

\subsection{Challenges}
\begin{itemize}
 \item Sample efficiency : Algorithms for handling varying environment conditions will definitely have issues w.r.t sample efficiency.
 When environment changes, then learning needs to be quick, but the speed will depend on the state-reward samples obtained. Hence, if these samples are not informative of the change, then algorithms might take longer to learn new policies from these samples. 
 \item Computation power : Single-agent RL algorithms face \emph{curse-of-dimensionality} with increased size of state-action spaces. Deep RL~\cite{book:drl} use graphical processing units (GPU) hardware for handling large problem size. Detecting changing operating conditions puts additional burden on computation. Hence, this will present a formidable challenge.
 \item Theoretical results : As stated in Section \ref{sec:prelim}, without Assumption \ref{as:statpr}, it is difficult to obtain convergence results for model-free RL algorithms in non-stationary environments. 
 Thus, providing any type of guarantees on their performance becomes hard.
\end{itemize}

\section{Current Solution Approaches}\label{sec:relatedwork}
Solution approaches proposed till now have looked at both finite 
horizon (see Section \ref{subsec:regretmin}) as well as infinite horizon (Section \ref{subsec:control}) cases. Prior approaches falling into these categories are described in the following subsections. 

\subsection{Finite Horizon Approaches}
\label{subsec:regretmin}
Finite horizon approaches to dealing with non-stationary environment are \cite{ucrl2}-\cite{acc2019}. 
These study MDPs with varying transition probability and reward functions. 
The performance criterion is the regret~\eqref{eqn:regretdef} and the goal of these algorithms is to minimize the regret over a finite horizon $T$. 
Since decision horizon is finite, the number of changes in the environment is utmost $T-1$.
Additionally, a stationary policy need not be optimal in this scenario. So, regret needs to be measured with respect to the best 
time-dependent policy starting from a state $s_0$. Basically, regret measures the sum of missed rewards when compared to the best policy (time-dependent) in hindsight.
\subsubsection{Comparison of the works}
How do the works \cite{ucrl2}-\cite{acc2019} compare with each other?
\begin{itemize}
\item All have similar objective - i.e., to minimize the regret during learning. Unlike infinite-horizon results which maximize the long-run objective and also provide methods to find optimal policy corresponding to this optimal objective value, regret-based learning approaches minimize regret during learning phase only. There are no known theoretical results to obtain a policy from this optimized regret value. Moreover, the regret value is based on the horizon length $T$.

\item The works \cite{ucrl2}-\cite{acc2019} slightly differ with regard to the assumptions on the pattern of environment changes.
\cite{ucrl2} assumes that the number of changes is known, while \cite{gajane,acc2019} do not impose restrictions on it.
The work on Contextual MDP~\cite{cmdp} assumes a finite, known number of environment contexts.
\cite{dick} assumes that only the cost functions change and that they vary arbitrarily.

\item Other than the mathematical tools used, the above works also differ with respect to the optimal time-dependent policy used in the computation of the regret. The optimal policy is average-reward optimal in \cite{ucrl2}, while it is total-reward optimal in \cite{dick}-\cite{acc2019}. \cite{cmdp} differs by letting the optimal policy to be a piecewise stationary policy, where each stationary policy is total-reward optimal for a particular environmental context.
\end{itemize}
\subsubsection{Details}
We now describe each of the above works in detail.
Contextual MDP (CMDP) is introduced by \cite{cmdp}. A CMDP is a tuple $\langle \mathcal{C} , S, A, Y(\mathcal{C})\rangle$, 
where $\mathcal{C}$ is the set of contexts and $c \in \mathcal{C}$ is the context variable. $Y(\mathcal{C})$ maps a context $c$ to a MDP $M_c = \langle S, A, P_c, R_c, \xi^c_0\rangle$. Here, $P_c$ and $R_c$ are same as $P_k$, $R_k$ respectively as defined in Section \ref{sec:probdef}. $\xi^c_0$ is the distribution of the initial state $s_0$. The time horizon $T$ is divided into $H$ episodes, with an MDP context $c \in \mathcal{C}$ picked at the start of each episode. This context chosen is latent information for the RL controller. 
After the context is picked (probably by an adversary), a start state $s_0$ is picked according to the distribution $\xi^c_0$ and episode sample path and rewards are obtained according to $P_c$, $R_c$. Suppose the episode variable is $h$ and $r_{h t}$ is the reward obtained in step $t$ of episode $h$. Let $T_h$, which is a stopping time, be the episode horizon. The regret is defined as follows:
\begin{equation*}
 \textnormal{Regret}_\textnormal{CMDP} = \sum\limits_{h=1}^{H} J^*_h - \sum\limits_{h=1}^{H} \sum\limits_{t=1}^{t_h} r_{h t},
\end{equation*}
where $J^*_h = J_h^{\pi^*_c} = \E{ \left[ \sum\limits_{t=0}^{T_h} r_{ht} | s_0 \sim \xi^c_0, \pi^*_c \right] }$ and $\pi^*_c$ is the optimal policy for context $c$. Note that $c$ is hidden and hence the above regret notion cannot be computed, but can only be estimated empirically.
The CECE algorithm proposed by \cite{cmdp} clusters each episode into one of the contexts in $\mathcal{C}$, based on the partial trajectory information. 
Depending on the cluster chosen, the context is explored and rewards are obtained. 

The UCRL2~\cite{ucrl2} and its improvement, variation-aware UCRL2~\cite{gajane} are model-based regret minimization algorithms, which  estimate the transition probability function as well as the reward function for an environment. These algorithms are based on the diameter information of MDPs, which is defined as follows:
\begin{equation}
\label{eqn:diam}
 D_{M} = \max_{s \neq s'} \min_{\pi:S \ra A} \E{[T(s'|s,\pi)]},
\end{equation}
where $M$ is the environment context and $T$ is the first time step in which $s'$ is reached from the initial state $s$.
Both algorithms keep track of the number of visits as well as the emprical average of rewards for all state-action pairs.
Using a confidence parameter, confidence intervals for these estimates are maintained and improved. The regret is defined as follows:
\begin{equation*}
 \textnormal{Regret}_\textnormal{UCRL2} = T\rho^* - \sum\limits_{t=1}^{T} r_t,
\end{equation*}
where $r_t$ is the reward obtained at every step $t$ and $\rho^*$ is the optimal average reward defined as follows:
\begin{equation*}
 \rho^* = \lim_{T \ra \infty} \frac{1}{T} \; \E{ \left[ \sum\limits_{t=1}^{T} r^*_t \right] },
\end{equation*}
$r^*_t$ is he reward obtained at every step when optimal policy $\pi^*$ is followed.
\\When environment model changes utmost $L$ times, then learning is restarted with a confidence parameter that is dependent on $L$.  Variation-aware UCRL2~\cite{gajane} modifies this restart schedule, where the confidence parameter is dependent on the MDP context variation also. Context variation parameter depends on the maximum difference between the single step rewards as well as the maximum difference between transition probability functions, over the time horizon $T$.
When the environment changes, the estimation restarts, leading to a loss in the information collected. Both algorithms give sublinear regret upper bound dependent on the diameter of the MDP contexts. The regret upper bound in \cite{gajane} is additionally dependent on the MDP context variation.

Online learning~\cite{oco} based approaches for non-stationary environments are proposed by \cite{dick,acc2019}. 
$\textnormal{MD}^2$~\cite{dick} assumes that the transition probabilities are stationary and known to the agent, while the cost functions vary (denoted $L_t$) and are picked by an adversary. The goal of the RL agent is to select a sequence of vectors $w_t \in \mathcal{C}_V$, where $\mathcal{C}_V \in \R^d$ is a convex and compact subset of $\R^d$. The chosen vectors must reduce the regret, which is defined as follows:
\begin{equation*}
 \textnormal{Regret}_\textnormal{MD2} = \sum_{t=1}^{T} \langle L_t,w_t\rangle - \min_{w \in \mathcal{C}_V} \sum_{t=1}^{T} \langle L_t,w\rangle,
\end{equation*}
where $\langle \cdot \,,\,\cdot \rangle$ is the usual Euclidean inner product. Thus, without information of $L_t$, $w_t$ can be chosen only by observing the history of cost samples obtained.
For this, the authors propose solution methods based on \emph{Mirror Descent} and \emph{exponential weights} algorithms.
\cite{acc2019} considers time-varying reward functions and develops a distance measure for reward functions, based on total variation.  Using this, regret upper bound is derived which depends on this distance measure. Further, \cite{acc2019} adapts \emph{Follow the Leader} algorithm for online learning in MDPs.

\subsubsection{Remarks} 
\begin{itemize}
\item Contextual MDP~\cite{cmdp} necessitates the need to measure ``closeness'' between MDPs, which enables 
the proposed CECE algorithm to cluster MDP models and classify any new model observed.
The clustering and classification of MDPs requires a distance metric for measuring how close are two trajectories to each other. \cite{cmdp} defines this distance using the transition probabilities of the MDPs. Using this distance metric and other theoretical assumptions, this work derives an upper bound on the regret, which is linear in $T$. The mathematical machinery used to show this is complex. Moreover, the distance measure used considers only the distance between probability distributions. However, the reward functions are important components of MDP and varies with the policy. It is imperative that a distance measure is dependent on reward functions too.

\item UCRL2 and variation-aware UCRL2 restart learning with changes in confidence parameter. This implies that in simple cases where the environment model alternates between two contexts, these methods restart with large confidence sets, leading to increased regret.
Even if this information is provided, these algorithms will necessarily require a number of iterations to improve the confidence sets for estimating transition probability and reward functions.

\item UCRL2, variation-aware UCRL2 and online learning approaches proposed in \cite{dick,acc2019} are model-based approaches, which do not scale well to large state-action space MDPs. The diameter $D_M$ (see \eqref{eqn:diam}) varies with the model and in many cases can be quite high, especially if the MDP problem size is huge. In this case, the regret upper bound might be very high.
\end{itemize}

\subsection{Infinite Horizon Approaches}
\label{subsec:control}
Works based on infinite-horizon are \cite{choi1}-\cite{nips2019}. These are oriented towards developing algorithms which learn a good control policy in non-stationary environment models. 
\subsubsection{Details}
\cite{choi1} proposes a stochastic model for MDPs with non-stationary environments. These are known as hidden-mode MDPs (HM-MDPs). Each mode corresponds to a MDP with stationary environment model. When a system is modeled as HM-MDP, then the transitions between modes are hidden from the learning agent.
State and action spaces are common to all modes - but each mode differs from the other modes w.r.t the transition probability and reward functions. Algorithm for solving~\cite{choi2} HM-MDP assumes that model information is known. 
It is based on a Bellman equation developed for HM-MDP which is further used to design a value iteration algorithm based on dynamic programming principles for this model.

A model-free algorithm for non-stationary environments is proposed by \cite{rlcd}. It is a context detection based method known as RLCD. Akin to UCRL2, RLCD estimates transition probability and reward functions from simulation samples. 
However, unlike UCRL2, it attempts to infer whether underlying MDP environment parameters have changed or not. The active model/context is tracked using a predictor function. This function utilizes an error score to rank the contexts that are already observed. The error score dictates which context is designated as ``active'', based on the observed trajectory. At every decision epoch, the error score of all contexts is computed and the one with the least error score is labeled as the current ``active'' model. 
A threshold value is used for the error score to instantiate data structures for new context, i.e., a context which is not yet observed by the learning agent. If all the contexts have an error score greater than this threshold value, then, 
data structures for a new context are initialized. This new context is then selected as the active context model. Thus, new model estimates and the associated data structures are created on-the-fly.

Suppose environment model changes are negligible, we expect that the value functions also do not change much amongst the models. This is formally shown by~\cite{csaji}. If the accumulated changes in transition probability or reward function remain \emph{bounded} over time and such changes are insignificant, then value functions of policies of all contexts are ``close'' enough. Hence, \cite{csaji} gives a theoretical framework highlighting conditions on context evolution. It also indicates when the pursuit for non-stationary RL algorithms is worthwhile. 

Change detection-based approaches for learning/planning in non-stationary RL is proposed by \cite{hadoux}-\cite{cql}.
The identification of active context based on the error score is the crux of RLCD method. \cite{hadoux} improves RLCD by
incorporating change detection techniques for identification of active context. Similar to RLCD, this method estimates the transition and reward functions for all contexts. Suppose the number of active context estimates maintained by \cite{hadoux} is $j$. At time $t$, a number $S_{i,t}$, $\forall \, i, \, 1 \leq i \leq j$ is computed. Let $\tilde{P}_i$ and $\tilde{R}_i$ be the transition probability and reward function estimates of context $i$, where $1 \leq i \leq j$. $S_{i,t}$ is updated as follows:
\begin{equation*}
 S_{i,t} = \max\left(0, S_{i,t-1} + \ln{ \frac{\tilde{P}_i(s_{t+1}|s_t,a_t) \tilde{R}_i(s_t,a_t,s_{t+1}) } {P_0(s_{t+1}|s_t,a_t) R_0(s_t,a_t,s_{t+1})}}\right),
\end{equation*}
where $P_0$ is the fixed transition function for a uniform model - one which gives equal probability of transition between all states for all actions and $R_0$ is set to $0$ for all state-action pairs. 
A change is detected if $\max\limits_{1\leq i \leq j} S_{i,t} > c$, where $c$ is a threshold value.
$\tilde{R}_i$ is updated as the moving average of simulated reward samples. $\tilde{P}_i$ is updated based on maximum likelihood estimation. The updation of $\tilde{P}_i$ and $\tilde{R}_i$ are same as in \cite{rlcd}.
\cite{qcd} shows that in full information case, i.e., when complete model information is known, the change detection approach of \cite{hadoux} leads to loss in performance with delayed detection. Based on this observation, with the full information assumption, \cite{qcd} designs a two-threshold policy switching method (denoted as TTS). 
Given the information that the environment switches from context $i$ to context $j$, TTS computes the Kullback-Leibler (KL) divergence of two contexts $P^{\pi_i}_i$ and $P^{\pi_i}_j$ w.r.t policy ${\pi_i}$, even though the policy ${\pi_i}$ is optimal for context $i$. 
When a sample tuple $(s_t,\pi_i(a_t),s_{t+1})$ comprising of current state, current action and next state is obtained at time $t$, the MDP controller computes the CUSUM~\cite{shiryaev} value $SR_t$ as follows:
\begin{equation}
\label{eqn:srt}
 SR_{t+1} = (1 + SR_t) \frac{P^{\pi_i}_j(s_{t+1} | s_t, {\pi_i}(a_t))}{P^{\pi_i}_i(s_{t+1} | s_t, {\pi_i}(a_t))}, \quad \; \; SR_0 = 0.
\end{equation}
If $SR_t$ is higher than a threshold value $c_1$, then it implies that the tuple $(s_t,\pi_i(a_t),s_{t+1})$ is highly likely to be originated in the context $j$, but it necessitates adequate exploration. 
Hence in every state, the action which maximizes the KL divergence between $P^{\pi_i}_j$ and $P^{\pi_i}_i$ is fixed as the exploring action. This policy is denoted as $\pi_{KL}$ and sample tuples starting from time $t+1$ are obtained using $\pi_{KL}$. 
Simultaneously $SR_t$ is also updated. When $SR_t$ crosses threshold $c_2$, where $c_2 > c_1$, TTS switches to $\pi_{j}$, which is the optimal policy for MDP with $P_j$ as the transition probability function.
The CUSUM statistic $SR_t$ helps in detecting changes in environment context.

\cite{cql} proposes a model-free RL method for handling non-stationary environments based on a novel change detection method for multivariate data~\cite{odcp}. Similar to \cite{qcd}, this work assumes that context change pattern is known. However, unlike \cite{qcd}, \cite{cql} carries out change detection on state-reward samples obtained during simulation and not on the transition probability functions. The Q-learning (QL) algorithm (see~\cite{sutton,ql}) is used for learning policies, but maintains a separate Q value table for each of the environment contexts. 
During learning, the state-reward samples, known as \emph{experience tuples}, are analyzed using the multivariate change detection method known as ODCP. When a change is detected, based on the known pattern of changes, the RL controller starts updating the Q values of the appropriate context. This method is known as \emph{Context QL} and is more efficient in learning in dynamically varying environments, when compared to QL.

A variant of QL, called as Repeated Update QL (RUQL) is proposed in \cite{ruql}. This adaptation of QL repeats the updates to the Q values of a state-action pair by altering the learning rate sequence of QL. Though this variant is simple to implement, it has the same disadvantage as QL, i.e., poor learning efficiency in non-stationary environments.

Online-learning based variant of QL for arbitrarily varying reward and transition probability functions in MDPs is proposed by~\cite{mannor2009}. This algorithm, known as Q-FPL, is model-free and requires the state-reward sample trajectories only. 
With this information, the objective of the algorithm is to control the MDP in a manner such that regret is minimized. 
The regret is defined as the difference between the average reward per step obtained by Q-FPL algorithm and the average reward obtained by the best stationary, deterministic policy. Formally, we have
\begin{equation*}
\begin{split}
 \textnormal{Regret}_\textnormal{Q-FPL} = \sup_{\sigma:S \ra A} \frac{1}{T} \sum\limits_{t=1}^{T} \E{ [r_t( s_t,\sigma(s_t) ) ] } - 
 \\ \frac{1}{T} \sum\limits_{t=1}^{T} \E{ [r_t( s_t,a_t) ) ] }
 \end{split}
\end{equation*}
where $r_1, r_2, \ldots$ are the arbitrary time-varying reward functions and $a_t$ is the action picked by Q-FPL. $\sigma$ is a stationary deterministic policy. Q-FPL partitions the learning iterations into intervals and in each interval, the Q values are learnt from the reward samples of that interval. These Q values are stored and are used to pick actions for the next interval by using the \emph{Follow the Perturbed Leader} strategy~\cite{oco}. At the end of every interval, Q values are reset to zero and not updated during the future intervals. The regret bounds for Q-FPL are derived by \cite{mannor2009}.

Similar to Q-FPL, the risk-averse-tree-search (RATS) algorithm \cite{nips2019} assumes minimal information regarding the evolution of environment conetxts. It requires slow evolution of environments, wherein information regarding current context is available to the RL  agent. RATS algorithm models dynamically varying context RL environments as a non-stationary MDP (NSMDP), which is a generalization of MDP. Thus, at every instant, a RL agent has access to the current ``snapshot'' of the environment in the NSMDP model. Given this snapshot and the current state, the RATS algorithm utilizes a tree-search algorithm to decide the optimal action to be exercised for the current state.

\subsubsection{Remarks}
\begin{itemize}
 \item The algorithms for solving HM-MDPs \cite{choi2} are computationally intensive and are not practically applicable. With advances in deep RL~\cite{book:drl}, there are better tools to make these computationally more feasible.
 
 \item RLCD~\cite{rlcd} does not require apriori knowledge about the number of environment contexts and the context change pattern, but is highly memory intensive, since it stores and updates estimates of transition probabilities and rewards corresponding to all detected contexts. 
 
 \item \cite{qcd} is a model-based algorithm and hence itis impossible to use it when model information cannot be obtained. However, this algorithm can be utilized in model-based RL. But certain practical issues limit its use even in model-based RL. One is that pattern of model changes needs to be known apriori. Additionally, its two-threshold switching strategy is dependent on CUSUM statistic for change detection and more importantly on the threshold values chosen. 
 Since \cite{qcd} does not provide a method to pre-fix suitable threshold values, it needs to be always selected by trial and error.
 This is impossible to do since it will depend on the reward values, sample paths etc.
 
 \item Extensive experiments while assessing the two threshold switching strategy put forth the following issue. This issue is with reference to \eqref{eqn:srt}, where the fraction $\frac{P^{\pi_i}_j(s_{t+1} | s_t, {\pi_i}(a_t))}{P^{\pi_i}_i(s_{t+1} | s_t, {\pi_i}(a_t))}$ is computed. Suppose for the policy $\pi_i$ it so happens that $P^{\pi_i}_j(s_{t+1} | s_t, {\pi_i}(a_t)) = P^{\pi_i}_i(s_{t+1} | s_t, {\pi_i}(a_t))$ and optimal policy of $P_j$ is $\pi_j \ne \pi_i$, we will have 
 $\frac{P^{\pi_i}_j(s_{t+1} | s_t, {\pi_i}(a_t))}{P^{\pi_i}_i(s_{t+1} | s_t, {\pi_i}(a_t))} = 1$ and $SR_t$ will grow uncontrollably and cross every pre-fixed threshold value. Thus, in this normal case itself, the detection fails, unless threshold value if pre-fixed with knowledge of the changepoint! Thus, \cite{qcd} is not practically applicable in many scenarios. 
 
 \item Numerical experiments in \cite{csaji} show that QL and asynchronous value iteration are adaptive in nature. So, even if environment contexts change, these learn the policies for the new context with the help of samples from the new context. However, once new samples are obtained and new context is sufficiently explored, the policies corresponding to older models are lost. Thus, QL does not have memory retaining capability. 
 
 \item RUQL~\cite{ruql} faces same issues as QL - it can learn optimal policies for only one environment model at a time and cannot retain the policies learnt earlier. This is mainly because both QL and RUQL update the same set of Q values, even if environment model changes. Further, QL and RUQL cannot monitor changes in context - this will require some additional tools as proposed by~\cite{cql}. The Context QL method retains the policies learnt earlier in the form of Q values for all contexts observed. This eliminates the need to re-learn a policy leading to better sample efficiency. This sample efficiency is however attained at the cost of memory requirement - Q values need to be stored for every context and hence the method is not scalable. 

\item RATS algorithm approximates a worst-case NSMDP. However, due to the planning algorithms required for the tree-search, this algorithm is not scalable to larger problems.
\end{itemize}
The prior approaches discussed in this section are summarized in Table \ref{table:summary}. The columns assess decision horizon, model information requirements, mathematical tools used and policy retaining capability. A `-' indicates that the column heading is not applicable to the algorithm.
In the next section, we describe works in related areas which are focussed on learning across different tasks or using experience gained in simple tasks to learn optimal control of more complex tasks. We also discuss how these are related to the problem we focus on.
{\renewcommand{\arraystretch}{1.2}
\begin{table*}[h]
    \begin{center}
        \begin{tabular}{|c|c|c|c|c|}
        \cline{1-5}
        Algorithm            &   Decision Horizon        & Model Information Requirements             &    Mathematical Tool Used  & Policy Retention\\        
        \cline{1-5}
        CECE~\cite{cmdp}      & Finite                   & Model-based                 &     Clustering and Classification  & -\\
        \cline{1-5}
        UCRL2~\cite{ucrl2}      & Finite                 & Model-based                 &     Confidence Sets  & - \\
        \cline{1-5}
        Variation-aware UCRL2~\cite{gajane} & Finite  & Model-based                 &     Confidence Sets  & - \\
        \cline{1-5}
        $\textnormal{MD}^2$~\cite{dick}  & Finite  & Partially Model-based                 &  Online learning  & - \\
        \cline{1-5}
        FTL~\cite{acc2019}      & Finite                 & Model-based                 & Online learning  & -\\
        \cline{1-5}
        RLCD~\cite{rlcd}      & Infinite                 & Model-free                 & Error score  & Yes \\
        \cline{1-5}
        TTS~\cite{qcd}      & Infinite                 & Model-based                 & Change detection  & - \\
        \cline{1-5}
        Context QL~\cite{cql}      & Infinite                 & Model-free                 &     Change detection  & Yes \\
        \cline{1-5}
        RUQL~\cite{ruql}      & Infinite                 & Model-free                 & Step-size manipulation  & No\\
        \cline{1-5}
        Q-FPL~\cite{mannor2009}      & Infinite                 & Model-free                 & Online learning  & Yes\\
        \cline{1-5}
        RATS~\cite{nips2019}      & Infinite                 & Model-free                 &     Decision tree search  & Yes \\
        \cline{1-5}
        \end{tabular}     
        \caption{\small Performance comparison of ODCP and ECP in changepoints detected when model information is known.}
        \label{table:summary}
       
    \end{center}
 \end{table*} 
}

\section{Related Areas}\label{sec:relatedareas}
\subsection{Continual Learning}
\label{subsec:continual}
Continual learning algorithms~\cite{clsurvey} have been explored in the context of deep neural networks. However, it is still in its nascent stage in RL. The goal in continual learning is to learn across multiple tasks. 
The tasks can probably vary in difficulty, but mostly they are the same problem domain. 
For e.g., consider a grid world task, wherein the RL agent must reach a goal position from a starting position by learning the movements possible, any forbidden regions etc. Note that the goal position matters in this task, since the agent learns to reach a given goal position. If the goal position changes, then it is a completely new task for the RL agent, which, now has to find the path to the new goal position. Thus, both tasks though being in the same problem domain are different. When the RL agent has to learn the optimal policy for the new grid world, it should make sure to not forget the policy for the old task. Hence, continual learning places emphasis on resisting forgetting~\cite{clsurvey}.

An agent capable of continual, hierarchical, incremental learning and development (CHILD) is proposed in~\cite{child}. This work introduces continual learning by stating the properties expected out of such a RL agent and combines temporal transition hierarchies (TTH) algorithm with QL. The TTH method is a constructive neural network based approach that predicts probabilities of events and creates new neuronal units to predict these events and their contexts. This method updates the weights, activations of the exising neuronal units and also creates new ones. It takes as input the reward signal obtained in the sample path. The output gives the Q values which are further utilized to pick actions. This work provides extensive experimental results on grid world problems where learning from previous experience is seen to outperform learning from scratch. The numerical experiments also analyze TTH's capability of acquiring new skills, as well as retaining learnt policies. 

\cite{cmplx_synapse} derives motivation from synaptic plasticity of human brain, which is the ability of the neurons in the brain to strengthen their connections with other neurons. These connections (or synapses) and strengths form the basis of learning in brain. Further, each of the neurons can simultaneously store multiple memories, which implies that synapses are capable of storing connection strengths for multiple tasks! \cite{cmplx_synapse} intends to replicate synaptic plasticity in neural network architectures used as function approximators in RL. For this, the authors use a biologically plausible synaptic model~\cite{bena-fusi}.
According to this model, the synaptic weight is dependent on a weighted average of its previous changes, which can further be approximated using a particular chain model. This chain model, which gives the synaptic weight at current time by accounting for all previous changes, is incorporated to tune the parameters of the neural networks. Experiments on simple grid world problems shows that QL with the above model has better performance in changing tasks when compared to classical QL.

Policy consolidation-based approach \cite{policy_consolidation} is developed to tackle forgetting of policies. It operates on the same synaptic model as \cite{cmplx_synapse}, but consolidates memory at the policy level. Policy consolidation means that the current behavioural policy is distilled into a cascade of hidden networks that record policies at multiple timescales. The recorded policies also affect the behavioural policy by feeding into the policy network. The policies are encoded by the parameters of the neural network and the distance between the parameters of two such networks can be used as a substitute for the distance between policies (represented by the networks). This substitute measure is also incorporated in the loss function used for training the policy network. This method is tested on some benchmark RL problems.

\subsubsection{Remarks}
\begin{itemize}
 \item Developing biologically inspired algorithms~\cite{cmplx_synapse,policy_consolidation} is a novel idea. This has been also explored in many areas in supervised learning as well. However, to develop robust performance which is reliable, adequate experimentation and theoretical justifications is needed. The above works lack this and at best can be considered as just initial advancements in this stream of work. 
 
 \item We would like to compare continual learning algorithms with approaches in Section \ref{sec:relatedwork}.  
 Algorithms like~\cite{gajane,ruql} do not resist \emph{catastrophic forgetting}, because training on new data quickly erases knowledge acquired from older data. These algorithms restart with a fixed confidence parameter schedule. In comparison to this, \cite{cql} adapts Q-learning for non-stationary environments. It resists catastrophic forgetting by maintaining separate Q values for each model. This work provides empirical evidence that policies for all models are retained. However, there are issues with computational efficiency and the method needs to be adapted for function-approximation based RL algorithms. 
 
 \item The CHILD~\cite{child} method is akin to RLCD~\cite{rlcd} and Context QL~\cite{cql}, both of which also have separate data structures for each model. Thus, in combination with change detection, the CHILD algorithm can be used for dynamically varying environments as well.
 
\end{itemize}

\subsection{Learning to Learn : Meta-learning Approaches}
\label{subsec:metalearn}

Meta-learning as defined in Section \ref{sec:intro} involves reusing experience and skills from earlier tasks to learn new skills. If RL agents must meta-learn, then we need to define what constitutes experience and what is the previous data that is useful in skill development. Thus, we need to understand what constitutes \emph{meta-data} and how to learn using meta-data. Most of the prior works are targeted towards deep reinforcement learning (DRL), where only deep neural network architectures are used for function approximation of value functions and policies.

A general model-agnostic meta-learning algorithm is proposed in~\cite{maml}. The algorithm can be applied to any learning problem and model that is trained using a gradient-descent procedure, but is mostly tested on deep neural architectures, since their parameters are trained by back propagating the gradients. The main idea is to get hold of an internal representation in these architectures that is suitable for a wide variety of tasks. Further, using samples from new tasks, this internal representation (in terms of network parameters) is fine-tuned for each task. Thus, there is no ``learning from scratch'', but learning from a basic internal representation is the main idea. The assumptions are that such representations are functions of some parameters and the loss function is differentiable w.r.t those parameters. The method is evaluated on classification, regression and RL benchmark problems.
However, it is observed by~\cite{tmaml} that the gradient estimates of MAML have high variance. This is mitigated by introducing surrogate objective functions which are unbiased.

A probabilistic view of MAML is given by~\cite{iclr2018}. A fixed number of trajectories from a task $T$ and according to a policy parameterized by $\theta$ is obtained. The loss function defined on these trajectories is then used to update the task-specific parameters $\phi$. This is carried out using the gradient of the loss function, which is obtained from either policy gradient~\cite{pg} or TRPO~\cite{trpo}. The same formulation is extended to non-stationary settings where \cite{iclr2018} assumes that tasks themselves evolve as a Markov chain.

Learning good directed exploration strategies via meta-learning is the focus in~\cite{mlexplore}. The algorithm developed by the authors, known as MAESN, uses prior experience in related tasks to initialize policies for new tasks and also to learn appropriate exploration strategies as well. This is in comparison to~\cite{maml}, where only policy is fine tuned.
The method assumes that neural network parameters are denoted $\theta$ and per-task variational parameters are denoted as $\omega_i$, $1 \leq i \leq N$, where $N$ is the number of tasks. On every iteration through the task training set, $N$ tasks are sampled from this training set according to a distribution $\rho$. For a task, the RL agent gets state and reward samples. These are used to update the variational parameters. Further, after the iteration, $\theta$ is updated using TRPO~\cite{trpo} algorithm.
Numerical experiments for MAESN are carried out on robotic manipulation and locomotion tasks.

\subsubsection{Remarks}
\begin{itemize}
 \item Explainability of generalization power of deep neural network architectures is still an open problem. The meta-RL approaches are all based on using these architectures. Thus, the performance of these algorithms can only be validated empirically. Also, most of the works described in this section lack theoretical justification. Only the problem formulation involves some mathematical ideas, but none of the results are theoretical in nature. However, applied works like~\cite{selfdrivingcars} can be encouraged, but only if such works provide some performance analysis of meta-RL algorithms.
 
 \item The experimental results in most of the above works are still preliminary. These can be improved by facilitating more analysis.
 
\end{itemize}

\section{Application Domains}\label{sec:appln}
Reinforcement learning finds its use in a number of domains - for e.g., in operations research~\cite{rl_in_or}, games~\cite{rl_in_games}, robotics~\cite{rl_in_robotics}, intelligent transportation and traffic systems~\cite{rl_in_transport}. 
However, in most of the prior works in these applications, the assumption is that the environment characteristics and dynamics remain stationary. 
The number of prior works developing application-specific non-stationary RL algorithms is limited. 
This is due to the fact that adapting RL to problems with stationary environments is the first simple step towards more general RL controllers, for which scalability is still an issue.
Only recent advances in deep RL~\cite{book:drl} has improved their scalability to large state-action space MDPs.
Improved computation power and hardware, due to advances in high-performance computing, has led to better off-the-shelf software packages for RL. Advancements in computing has led to better implementations of RL - these use deep neural architectures~\cite{book:drl}, parallelization~\cite{ray,rllib} for making algorithms scalable to large problem sizes. Single-agent RL algorithms are now being deployed in a variety of applications owing to the improved computing infrastructure. 
One would also expect that easing of the stationary assumptions on RL environment models would also further increase the need for high computation power. But, due to these advances in computing infrastructure, there is hope to extend RL to applications where non-stationary settings can make the system inefficient (unless there is adaptation).

In this section, we survey the following representative application domains: transportation and traffic systems, cyber-physical systems, digital marketing and inventory pricing, recommender systems and robotics. In these representative domains, we cover works which propose algorithms to specifically deal with dynamically varying environments. Most of these prior works are customized to their respective applications.

\subsection{Transportation and Traffic Systems}
Traffic systems are either semi-automated or fully automated physical infrastructure systems that manage the vehicular traffic in urban areas. These are installed to improve flow of vehicular traffic and relieve congestion on urban roads. With the resurgence of AI, these systems are being fully automated using AI techniques. AI-based autonomous traffic systems use computer vision, data analytics and machine learning techniques for their operation. Improvement in computing power for RL has catapulted its use in traffic systems, and, RL based traffic signal controllers are being designed~\cite{prabu-tsc,soilse,cql}. Non-stationary RL traffic controllers are proposed by~\cite{soilse,cql}.

\emph{Soilse}~\cite{soilse} is a RL-based intelligent urban traffic signal controller (UTC) tailored to fluctuating vehicular traffic patterns. It is phase-based, wherein a phase is a set of traffic lanes on which vehicles are allowed to proceed at any given time. Along with the reward feedback signal, the UTC obtains a degree of pattern change signal from a pattern change detection (PCD) module. This module tracks the incoming traffic count of lanes at a traffic junction. It detects a change in the flow pattern using moving average filters and CUSUM~\cite{cusum} tool. When a significant change in traffic density is detected, learning rates and exploration parameters are changed to facilitate learning.
Context QL~\cite{cql}, as described in Section \ref{sec:relatedwork}, tackles non-stationary environments. This method is evaluated in an autonomous UTC application. 
The difference in the performance of QL~\cite{ql} and Context QL is highlighted by numerical experiments~\cite{cql}. 
This difference indicates that designing new methods for varying operating conditions is indeed beneficial.

Intelligent transportation systems (ITS) employ information and communication technologies for road transport infrastructure, mobility management and for interfaces with other modes of transport as well. This field of research also includes new business models for smart transportation. Urban aerial transport devices like unmanned aerial systems (UAS) are also part of ITS.
For urban services like package delivery, law enforcement and outdoor survey, an UAS is equipped with cameras and other sensors. To carry out these tasks efficiently, UAS takes photos and videos of densely human populated areas. Though information gathering is vital, there are high chances that the UAS intrudes privacy.~\cite{uas} considers this conflicting privacy-information criteria. The problem is that UAS may fly over areas which are densely populated and take pictures of humans in various locations. Though the UAS can use human density datasets to avoid densely populated locations, human concentration is still unpredictable and may change depending on weather, time of day, events etc. Thus, the model of human population density tends to be non-stationary.~\cite{uas} proposes a model-based RL path planning problem that maintains and learns a separate model for each distribution of human density. 
%

\subsection{Cyber-Physical Systems and Wireless Networks}
Cyber-physical systems (CPS) are integration of physical processes and their associated networking and computation systems. A physical process for e.g., a manufacturing plant or energy plant, is controlled by embedded computers and networks, using a closed feedback loop, where physical processes affect computation and vice-versa. Thus, autonomous control forms an innate part of CPS. 
Many prior works address anomaly detection in CPS~\cite{sensorapp}, since abnormal operation of CPS forces the controllers to deal with non-stationary environments. 

CPS security~\cite{cps} also arises from anomaly detection. The computation and networking systems of CPS are liable to denial of service (DoS) and malware attacks. These attacks can be unearthed only if sensors and/or CPS controller can detect anomalies in CPS operation. In this respect,~\cite{cps} proposes a statistical method for operational CPS security. A modification of the Shiryaev-Roberts-Pollak procedure is used to detect changes in operation variables of CPS which can detect DDoS and malware attacks.

The data from urban infrastructure CCTV networks can be leveraged to monitor and detect events like fire hazards in buildings, organized marathons on urban roads, crime hot-spots etc.~\cite{cctv} uses CCTV data along with social media posts data to detect events in an urban environment. This multimodal dataset exhibits change in properties before, after and during the event. Specifically, \cite{cctv} tracks the counts of social media posts from locations in the vicinity of a geographical area, counts of persons and cars on the road. These counts are modeled as Poisson random variables and it is assumed that before, after and during a running marathon event, the mean rates of the observed counts changes. A hidden Markov model (HMM) is proposed with the mean count rates as the hidden states. This HMM is extended to stopping time POMDP and structure of optimal policies for this event detection model is obtained.

\cite{nonstatisit2020} considers improving user experience in cellular wireless networks by minimizing Age of Information metric (AoI).
This metric measures the freshness of information that is transmitted to end users (``user equipments'') in a wireless cellular network. A multi-user scheduling problem is formulated which does not restrict the characteristics of the wireless channel model. Thus, a non-stationary channel model is assumed for the multi-user scheduling problem and the objective is to minimize transmission of stale information from the base stations to the user equipments. For this, an infinite-state, average-reward MDP is formulated. Optimizing this MDP is infeasible and hence this work finds a simple heuristic scheduling policy which is capable of achieving the lowest AoI.

\subsection{Digital Marketing and Inventory Pricing}
Digital marketing and inventory pricing are connected strategies for improving sale of goods and services. In current times, many online sites complement inventory pricing with digital marketing to attract more buyers and hence improve sales. Digital marketing moves away from conventional marketing strategies in the sense that it uses digital technologies like social media, websites, display advertising, etc to promote products and attract customers. Thus, it has more avenues for an improved reach when compared to conventional marketing. Inventory pricing is concerned with pricing the goods/services that are produced/developed to be sold. It is important that to gain profits, the manufacturer prices products/services according to the uncertain demand for the product, the production rate etc. 

A pricing policy for maximizing revenue for a given inventory of items is the focus of~\cite{pricing}. The objective of the automated pricing agent is to sell a given inventory before a fixed time and maximize the total revenue from that inventory. This work assumes 
that the demand distribution is unknown and varies with time. Hence, this gives rise to non-stationary environment dynamics. This work employs QL with eligibility traces~\cite{sutton} to learn a pricing policy.

~\cite{digital_mktg} studies off-policy policy evaluation method for digital marketing. The users of an online product site are shown customized promotions. Every such marketing promotion strategy uses the customer information to decide which promotions to display on the website. \cite{digital_mktg} proposes a MDP model with user information as the state and the promotions to be displayed as the action. The reward gained from promotions is measured by tracking the number of orders per visit of the customer. 
The proposed method is shown to reduce errors in policy evaluation of the promotion strategy.

\subsection{Recommender Systems}
Recommender systems/platforms are information filtering systems that predicts the preferences that a user would assign to a product/service. These systems have revolutionized online marketing, online shopping and online question-answer forums etc.
Their predictions are further aimed at suggesting relevant products, movies, books etc to online users. These systems now form the backbone of digital marketing and promotion. Many content providers like Netflix, YouTube, Spotify, Quora etc use them as content recommenders/playlists.

A concept drift based model management for recommender systems is proposed by~\cite{recommender1}. This work utilizes RL for handling concept drift in supervised learning tasks. Supervised learning tasks see shifts in input-label correspondence, feature distribution due to ever changing dynamics of data in real world. Each feature distribution and input-label correspondence is represented as a model  and whenver there is a shift in the underlying data, this model needs to be retrained. A MDP is formulated for taking decisions about model retraining, which decides when to update a model. This decision is necessary, because, the model of a given system influences the ability to act upon the current data and any change in it will affect its influence on current as well as future data. If new models are learned quickly, then the learning agent may be simply underfitting data and wasting computational resources on training frequently. However, if the agent delays model retraining, then the prediction performance of model might decrease drastically. Thus, given the current model, current data, the MDP-based RL agent decides when and how to update the model. A similar work using variants of deep Q-networks (DQN)~\cite{dqn} is proposed in~\cite{recommender2}.

\subsection{Robotics}
Robotics is the design, development, testing, operation and the use of robots. Its objective is to build machines that are intelligent,  can assist humans in various tasks and also perform tasks which are beyond human reach. Robots can also be designed to provide safety to human operations. 
Robots are now being utilized in outer space missions, medical surgery, meal delivery in hospitals~\cite{corona} etc. However, often robots need to adapt to non-stationary operating conditions - for e.g., a ground robot/rover must adapt its walking gait to changing terrain conditions~\cite{humanoid} or friction coefficients of surface~\cite{robot1}. 

Robotic environments characterized by changing conditions and sparse rewards are particularly hard to learn because, often, the reinforcement to the RL agent is a small value and is also obtained at the end of the task. \cite{robot1} focuses on learning in robotic arms where object manipulation is characterized by sparse-reward environments. The robotic arm is tasked with moving or manipulating objects which are placed at fixed positions on a table. In these tasks, often, dynamic adaptation to the surface friction and changed placement of objects on the table is tough.~\cite{robot1} adapts the TRPO algorithm for dealing with these changing operating conditions. The robotic arm RL agent is modeled as a continuous state-action space MDP. 
In a continuous state-action space setting, the policy is parameterized by Gaussian distribution.~\cite{robot1} proposes a strategy to adjust the variance of this Gaussian policy in order to adapt to environment changes.

Hexapod locomotion in complex terrain is the focus of~\cite{hexapod}. This approach assumes that the terrain is modeled using $N$ discrete distributions and each such distribution captures the difficulties of that terrain. For each such terrain, an expert policy is obtained using deep RL. Conditioned on the state history, a policy from this set of expert policies is picked leading to an adaptive gait of hexapod.

\subsection{Remarks}
All prior works discussed in this section are specifically designed for their respective applications. For e.g., Soilse~\cite{soilse} predicts the change in lane inflow rates and uses this to infer whether environment context has changes or not. This technique is limited to the traffic model and more so if lane occupation levels are the states of the model. Similar is the case with a majority of the other works as well. It is tough to extend the above works to more general settings. Some works which are generalizable are \cite{cql,uas,digital_mktg,robot1,hexapod}. 
The methods suggested in these works can be adapted to other applications as well provided some changes are incorporated. For e.g., \cite{cql} should be extended to continuous state-action space settings by incorporating function approximation techniques. This will improve its application to tougher problems. \cite{uas} utilizes Gaussian process tool to build a model-based RL path planner for UAS.
This can be extended to model-free settings using \cite{gp1} or other works on similar lines. Extending \cite{digital_mktg} to policy improvement techniques like actor-critic~\cite{ac} and policy gradient~\cite{pg} is also a good direction of future work.

\section{Future Directions}\label{sec:fw}
The previous sections of this survey introduced the problem, presented the benefits and challenges of non-stationary RL algorithms as well as introduced prior works. This survey paper also categorized earlier works. In this section, we describe the possible directions in which the prior works can be enhanced. Following this, we also enumerate challenges which are not addressed by the prior works, and which warrant our attention.

Prior approaches can be improved in the following manner:
\begin{itemize}
    \item The regret based approaches described in Section \ref{subsec:regretmin} are useful in multi-armed bandit-type settings where efficient learning with minimal loss is the focus. Since these are not geared towards finding good policies, these works do not prove to be directly useful in RL settings, where control is the main focus. However, the ideas they propose can be incorporated to guide initial exploration of actions in approaches like \cite{cql,ruql}. 
    
    \item Relaxing certain theoretical assumptions like non-communicating MDPs~\cite{fruit}, multi-chain MDPs~\cite{multichain} etc can further improve the applicability of regret-based approaches in control-based approaches. 
    
    \item Most of the model-based and model-free approaches in Section \ref{sec:relatedwork} are not scalable to large problem sizes. This is because each of these methods either consume lot of memory for storing estimates of model information~\cite{ucrl2}-\cite{hadoux}, or consume compute power for detecting changes \cite{qcd,cql}. \cite{nips2019} uses compute power for building large decision trees as well. These phenomenal compute power and memory requirements render these approaches to be non-applicable in practical applications which typically function with restricted resources. An option is to offload the compute and memory power requirements onto a central server.
    Another option is to incorporate function approximation in the representation of value functions and policies.
      
    \item Tools from statistics - like for e.g., quickest change detection~\cite{shiryaev}, anomaly detection can prove to be indispensable in the problem of non-stationary RL. Also introducing memory retaining capacity in deep neural network architectures will can be a remedy for resisting catastrophic forgetting.
    
    \item Works~\cite{qcd,cql,soilse} assume that the pattern of environment changes is known and can be tracked. However, practically it is often difficult to track such changes. For this tracking~\cite{satrack} methods can be used.
\end{itemize}
Next, we discuss additional challenges in this area.
\begin{itemize}
    \item Need to develop algorithms that are sensitive to changes in environment dynamics and adapt to changing operating conditions seamlessly. Such algorithms can be extended to continual RL settings.
    
    \item In the literature, there is a lack of Deep RL approaches to handle non-stationary environments, which can scale with the problem size. Meta learning approaches~\cite{maml,iclr2018,mlexplore,tmaml} exist, but these are still in the initial research stages. These works are not sufficiently analyzed and utilized. More importantly, these are not explainable algorithms. 
    
    \item Some applications like robotics~\cite{rl_in_robotics} create additional desired capabilities like for e.g \emph{sample efficiency}. When dealing with non-stationary environment characteristics, the number of samples the RL agent obtains for every environment model can be quite limited. In the extreme case, the agent may obtain only one sample trajectory, which is observed in robotics arm manipulation exercises. In such a case, we expect the learning algorithm to be data efficient and utilize the available data for multiple purposes - like learn good policies as well as detect changes in environment statistics.
    
    \item While encountering abnormal conditions, a RL autonomous agent might violate safety constraints, because the delay in efficiently controlling the system in abnormal conditions can lead to some physical harm. For e.g., in self-driving cars, a suddenor abrupt change in weather conditions can lead to impaired visual information from car sensors. Such scenarios mandate that the RL agent, though still learning new policies, must keep up with some nominal safe bahaviour. Thus, this can lead to works which intersect safe RL~\cite{saferl} and non-stationary RL algorithms.
    
\end{itemize}

\bibliographystyle{IEEEtran}
\bibliography{survey_nonstatrl_ref}

\end{document}